\documentclass[10pt,a4paper, conference]{IEEEtran}
\IEEEoverridecommandlockouts
\usepackage{cite}
\usepackage{amsmath,amssymb,amsfonts}
\usepackage{algorithmic}
\usepackage{graphicx}
\usepackage{textcomp}
\usepackage{xcolor}
\def\BibTeX{{\rm B\kern-.05em{\sc i\kern-.025em b}\kern-.08em
    T\kern-.1667em\lower.7ex\hbox{E}\kern-.125emX}}

\usepackage{multirow}
\usepackage{caption}
\usepackage{subcaption}
\usepackage{booktabs}
\usepackage{tabularx}
\usepackage{url}
\usepackage{bbm}

\newcommand{\ie}{\textit{ie.}}

\def\BibTeX{{\rm B\kern-.05em{\sc i\kern-.025em b}\kern-.08em
    T\kern-.1667em\lower.7ex\hbox{E}\kern-.125emX}}
\begin{document}
%

\title{\emph{DR$^2$S}: Deep Regression with Region Selection for Camera Quality Evaluation}

\author{\IEEEauthorblockN{Marcelin Tworski}
\IEEEauthorblockA{\textit{LTCI, T\'{e}l\'{e}com Paris, }\\
\textit{Institut polytechnique de Paris}\\
Palaiseau, France \\
and \textit{DXOMARK}\\
Boulogne-Billancourt, France \\
marcelin.tworski@telecom-paris.fr}
\and
\IEEEauthorblockN{St\'{e}phane Lathuili\`{e}re}
\IEEEauthorblockA{\textit{LTCI, T\'{e}l\'{e}com Paris, }\\
\textit{Institut polytechnique de Paris}\\
Palaiseau, France \\
stephane.lathuiliere@telecom-paris.fr}
\and
\IEEEauthorblockN{Salim Belkarfa}
\IEEEauthorblockA{\textit{DXOMARK} \\
Boulogne-Billancourt, France \\
sbelkarfa@dxomark.com}
\and
\IEEEauthorblockN{\hspace{3.8cm}Attilio Fiandrotti}
\IEEEauthorblockA{\hspace{3.8cm}\textit{LTCI, T\'{e}l\'{e}com Paris, }\\
\textit{\hspace{3.8cm}Institut polytechnique de Paris}\\
\hspace{3.8cm}Palaiseau, France \\
\hspace{3.8cm}attilio.fiandrotti@telecom-paris.fr}
\and
\IEEEauthorblockN{Marco Cagnazzo}
\IEEEauthorblockA{\textit{LTCI, T\'{e}l\'{e}com Paris, }\\
\textit{Institut polytechnique de Paris}\\
Palaiseau, France \\
marco.cagnazzo@telecom-paris.fr}
}


%



\maketitle
\begin{abstract}
In this work, we tackle the problem of estimating a camera capability to preserve fine texture details at a given lighting condition. Importantly, our texture preservation measurement should coincide with human perception. 
Consequently, we formulate our problem as a regression one and we introduce a deep convolutional network to estimate texture quality score. At training time, we use ground-truth quality scores provided by expert human annotators in order to obtain a subjective quality measure.
In addition, we propose a region selection method to identify the image regions that are better suited at measuring perceptual quality. Finally, our experimental evaluation shows that our learning-based approach outperforms existing methods and that our region selection algorithm consistently improves the quality estimation.

%
\end{abstract}

%
\IEEEpeerreviewmaketitle

\section{Introduction}

With the rapid rise of smartphone photography, many people got more interested in their photographs and in capturing high-quality images. Smartphone makers responded by putting an increasing emphasis on improving their cameras and image processing systems. In particular, sophisticated imaging pipelines are now used in the Image Signal Processor (ISP) of any smartphone. Such software makes use of recent advances such as multi-camera fusion \cite{trinidad2019multi}, HDR+ \cite{hasinoff2016burst} or deep learning-based processing to improve image quality. Despite the high performance of those techniques, their increasing complexity leads to a correspondingly increased need for quality measurement.

Several attributes are important to evaluate an image: target exposure and dynamic range, color (saturation and white balance) texture, noise and various artifacts that can affect the quality of the final image \cite{vcadik2006image}. When it comes to camera evaluation,  a camera can be evaluated for its capabilities in low-light,  zoom, or shallow depth-of-field simulation.

In this work, we are interested in estimating the quality of a camera for multiple lighting conditions. More specifically, we aim at evaluating camera capabilities to preserve fine texture details. We also refer to this problem as texture quality estimation. Texture details preservation should be measured in a way that reflects human perception. 
A typical way to evaluate the quality of a set of cameras consists of comparing shots of the same visual content in a controlled environment. The common visual content is usually referred to as a \emph{chart}. The motivation for using the same chart when comparing different cameras is twofold. First, it facilitates the direct comparison of different cameras. In particular, when it comes to subjective evaluation, humans can more easily provide pairwise preferences than an absolute quality score. Second, when the common noise-free chart is known, this reference can be explicitly included in the quality measurement process.
In this context, the Modulation Transfer Function (MTF) is a widely-used tool for evaluating the perceived sharpness and resolution, which are essential dimensions of texture quality. First, MTF-based methods suffer from important drawbacks. MTF-based methods are originally designed for conventional optic systems that can be modeled as linear. Consequently, non-linear processing in the camera processing pipeline, such as multi-images fusion or deep learning-based image enhancement, may lead to inaccurate quality evaluation~\cite{van2019edge}. 

Second, these methods assume that the norm of the device transfer function is a reliable measure of texture quality. However, in this work, we argue that this assumption fails to account for many nuances of human perception. Some recent works have shown that the magnitude of image transformations do not always coincide with the perceived impact of the transformations~\cite{zhang2018unreasonable}. Consequently, we advocate that human judgment should more explicitly be included in the texture quality measurement process.

As a consequence, we propose \emph{DR$^2$S}, a Deep Regression method with Region Selection. Our contributions are threefold. First, we formulate the problem of assessing the texture quality of a device as a regression problem and we propose to use a deep convolutional network (ConvNet) for estimating quality scores. We aim at obtaining a score which would be close to a subjective quality judgment: to this end, we use annotations provided by expert human observers as ground-truth at training time. Second, we propose an algorithm to identify the regions in a chart that are better suited to measure perceptual quality. Finally, we perform an extensive evaluation study that shows that our learning-based approach performs better than existing methods for texture quality evaluation and that our region selection algorithm leads to further improvement in texture quality measurement.
\section{Related Works}

In this section, we review existing works on texture quality assessment. Existing methods can be classified into two main categories: MTF-based and learning-based methods. 

\subsection{MTF-based methods}\label{DLAcutanceDesc}

A simple model for a camera consists in a linear system that produces an image $y$ as a convolution of the point spread function $h$ and the incoming radiant flux $x$. In the frequency domain, $Y(f) = H(f)X(f) + N(f)$, where we also consider additive noise $N$. The
modulation transfer function is $\operatorname{MTF}(f) = |H(f)|$ and it is commonly used to characterize an optic acquisition device ~\cite{boreman2001modulation}.

Acutance is a single-value metric calculated from the device MTF. 
To compute this value, a contrast sensitivity function (CSF), modeling the spatial response of the human visual system, is used to weight the values of the MTF for the different spatial frequencies. The CSF depends on a set of parameters named viewing conditions, which describes how an observer would look at the resulting image. These parameters are usually the printing height and the viewing distance. The acutance is defined as the ratio of the integral of the weighted MTF by CSF's integral.

The key assumption in MTF-based methods is that they dispose of the noise-free content to estimate the transfer function. The noise-free content is often referred to as \emph{full reference}.  Therefore, these methods are usually used only with synthetic visual charts. To the best of our knowledge, the only work estimating acutance on natural scenes was proposed by Van Zwanenberg et al. \cite{van2019edge}. This method uses edges detection in the scene to compute the MTF. Early methods use charts containing a blur spot or a slanted edge for this computation. Loebich et al. ~\cite{loebich2007digital} propose a method using the Siemens-Star. Cao et al. propose to use the Dead-Leaves model ~\cite{matheron1975random, gousseau2007modeling}, and introduce an associated method in~\cite{cao2009measuring}, which is shown to be more appropriate to describe fine detail rendering since its texture are more challenging for devices. This chart is usually referred to as the \emph{Dead-Leaves} chart. In this chart, the reference image is generated with occluding disks generated with a random center location, radius and grey-scale value.
Importantly, digital camera systems present high-frequency noise, which affects the MTF estimation by dominating signal in the higher frequencies. Consequently, estimating the noise power spectral density (PSD) is key to obtain an accurate acutance evaluation, and this task is not easily performed on the textured region. As a consequence, noise PSD is typically estimated on a uniform patch. It is important to note that only the PSD of the reference image and not the reference image itself is needed for the acutance computation. For this reason, this method is referred to as Reduced-Reference (RR) acutance in the rest of this paper. 
However, this approach is hindered by the denoising algorithms integrated into cameras. Not only these algorithms interfere with the noise PSD estimation but also they behave differently in uniform and textured regions. In this context, Kirk et al. \cite{kirk2014description} propose to compute the MTF using the cross-power spectral density of the resulting and the reference images. This method assumes an effective registration of the chart. Sumner et al. \cite{sumner2017effects} then modified Kirk's computation in order to make it more robust to slight misregistration. Since this method takes fully advantage of the reference image, it is referred to as Full-Reference (FR) acutance in the rest of this paper.

In conclusion, state-of-the-art techniques typically allow to obtain a good estimation of devices' MTF and then of the acutance. However, it has been shown that acutance itself does not always reflect very well the human quality judgment. This observation calls for learning-based methods that aim at reproducing the score of human experts evaluating the images.

\subsection{Learning-based methods}

In opposition to MTF-based techniques described in the previous section, learning-based methods require annotated datasets. Early datasets (LIVE \cite{sheikh2006statistical}, CSIQ \cite{larson2010most}, TID2008 \cite{ponomarenko2009tid2008} and its extension TID2013 \cite{ponomarenko2015image}) consist of noise-free images and subjective preference scores for different types and levels of artificially introduced distortions.
These distortions mostly correspond to compression or transmission scenarios.  Only some of them, such as intensity shift, additive Gaussian noise, contrast changes or chromatic aberrations, are relevant to the problem of camera evaluation, and those distortions do not encompass the complexity of real systems. Conversely, the KonIQ10k \cite{hosu2020koniq} dataset consists of samples from a larger public media database with unknown distortions claimed to be authentic. Similarly, the LIVE In the Wild \cite{7327186} database consists of 1161 images shot with various devices with unique content for the images. For both datasets, crowd-sourced studies were conducted in order to collect opinions about the quality of these images. Interestingly, the size of these datasets is sufficiently large to allow the use of deep learning approaches. In particular, Varga et al. \cite{varga2018deeprn} propose to fine-tune a ResNet-101 network \cite{he2016deep} equipped of spatial pyramid pooling layers \cite{he2015spatial} in order to accept the full images at a non-fixed size to predict the user preference score. This deep approach leads to the best performances on these authentic distortions benchmarks. 

Despite the good results of this method, it cannot be directly applied to our problem. One of the reasons is that we aim at estimating the quality of the full image produced by the camera: resizing it would alter the fine details we try to estimate, and without resizing we undergo memory usage issues. To tackle this problem, we propose to work on patches randomly cropped in the input image.

More recently, Yu et al. \cite{yu2018learning} collected a dataset of 12 853 natural photos from Flickr and used Amazon Mechanical Turk to rate them according to image quality defects: exposition, white balance, color saturation, noise, haze, undesired blur, composition. They used a multi-column CNN: Using GoogLeNet \cite{szegedy2015going}, the first layers are shared, and then it is separated according to each attribute. For learning efficiency purposes, they frame the problem of each defect regression as an 11-class classification task.

Nevertheless, these deep learning approaches \cite{he2015spatial, yu2018learning} tackle a slightly different problem than ours. They are designed to evaluate image quality attributes for any input image while we address the problem of evaluating devices using a known common chart. 


To address the limitations of both MTF-based and learning-based methods, we now introduce our novel deep regression framework for texture quality estimation.


\begin{figure*}[t]
\centering
\includegraphics[width = 0.8\textwidth]{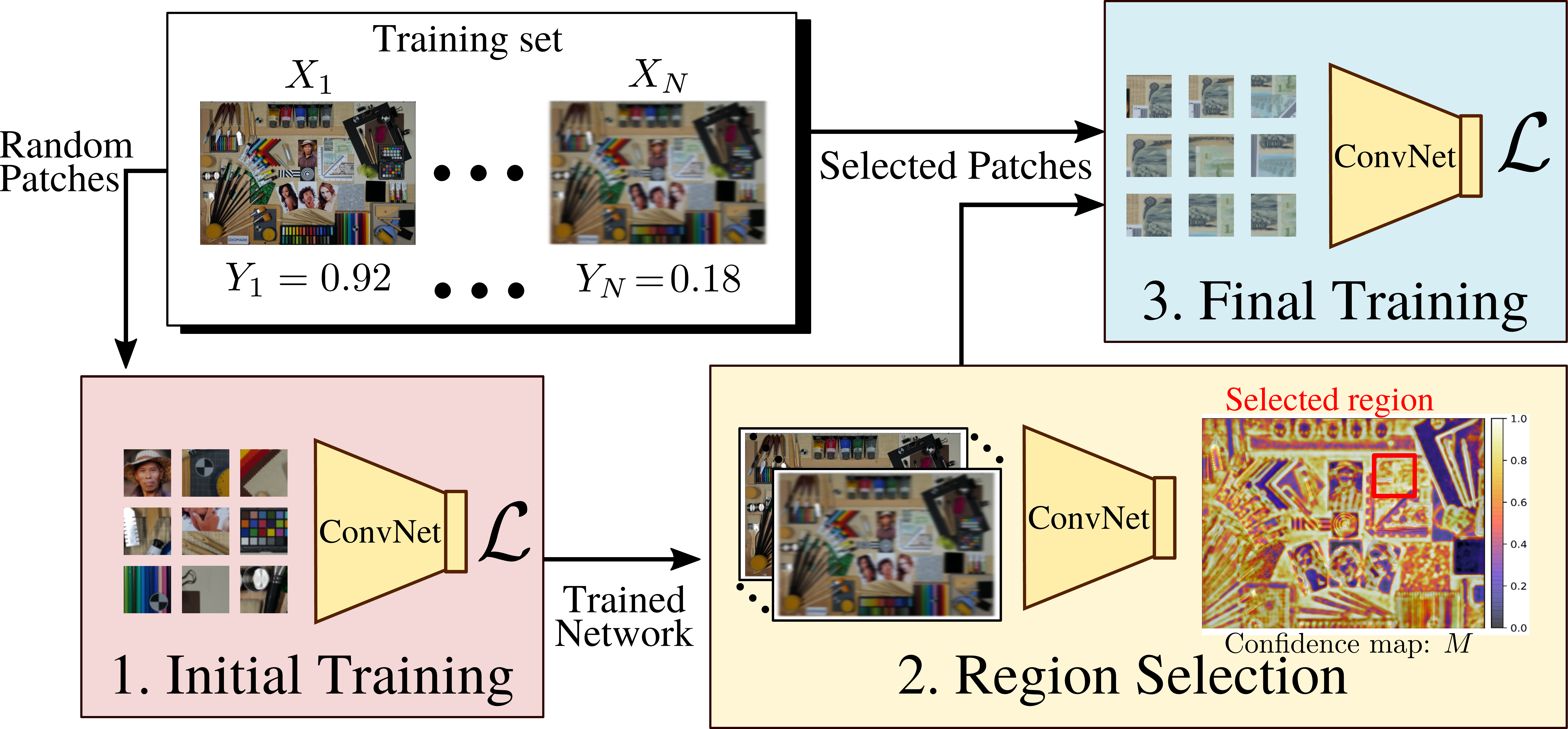}
\caption{Pipeline overview of DR$^2$S: Our method is divided into three main stages. First, we perform a naive training using patches randomly cropped over the chart. Second, we compute a map that indicates discriminant regions. Third, we perform a final training using only a selected region.}
\label{fig:pipeline}
\end{figure*}

\section{Method}
\label{sec:method}
\subsection{DR$^2$S Overview}
In this section, we detail the proposed method for estimating texture quality. We formulate this task as a regression problem. We assume that we dispose of a training dataset composed of N color images $(X_1,\hdots,X_N)$ of dimensions $H\times W$ with the corresponding texture quality scores $(Y_1,\hdots,Y_N) \in \mathbb{R}^N$.
Since we are interested in estimating the camera quality, each score $Y_n$ corresponds to a quality score for one device at a specific lighting condition. Note that, several training images can be taken with the same device. 
Importantly, we aim at computing quality scores that coincide with human judgment, the ground-truth texture quality scores are provided by human annotators (See Sec.\ref{sec:datasets} for more details). 
We aim at training a ConvNet $\phi(\cdot,\theta)$ with parameters $\theta$. We consider that these images depict the same chart and are taken with different cameras and different lighting conditions. 


Our method is based on the observation that all image regions are not equally suited to predict the overall image quality. For example, regions with uniform textures will be rendered similarly by any device independently of its quality. Conversely, other regions with rich and fine texture details are much more discriminating since they will be differently captured by different devices. Based on this observation, we propose the algorithm illustrated in Fig.~\ref{fig:pipeline}. The method is divided into three stages: First, we train a first neural network ignoring this problem of unsuited regions. Second, we employ this network for estimating which image regions are the most suitable for measuring image quality. Finally, the network is re-trained using only selected regions. In the following we provide the details of the three stages of our algorithm.

\subsection{Initial Training}
The goal of this first stage is to train a neural network that can be used to identify relevant image regions. To this end, we propose to train a network to regress the quality score from an input image patch. This initial network is later used in the second stage of our pipeline to identify discriminant regions (See Sec.\ref{sec:map}). We train this deep convNet on random crops extracted from the training images $\{X_1,\hdots,X_N\}$. These crops are randomly selected across the images with a uniform distribution.  
In all our experiments, we use the widely used Resnet-50 network pre-trained on ImageNet \cite{deng2009imagenet} where the final classification layer is replaced by a linear fully connected layer. Since some patches are not discriminant, this initial training suffers from instability and optimization issues. Consequently, we use the Huber loss that reports better performance in the presence of noisy samples \cite{carvalho2018regression}:\begin{equation}
\mathcal{L}(y, \hat{y})=\left\{\begin{array}{ll}
\frac{1}{2}(y-\hat{y})^{2}  &\text { for }|y-\hat{y}| \leq \delta \\
\delta|y-\hat{y}|-\frac{1}{2} \delta^{2}  & \text { otherwise }
\end{array}\right.
\end{equation}
\noindent
where $y$ and $\hat{y}$ denote the annotated and predicted scores, and $\delta\in\mathbb{R^+}$ is a threshold.
At the end of this stage, we obtain a network that estimates the quality of a device from an input patch.

\subsection{Region Selection}
\label{sec:map}
In the second stage of our pipeline, we use our previously trained network to predict image regions that are most discriminating for quality measurement. We produce a map $M\in[0,1]^{H\times W}$ that indicates the relevance of each location of the chart to estimate texture quality. 
This map will allow us to select a suitable region to train our convNet in the last stage of our pipeline.

In order to estimate a single map $M$ for all the training images, we first register all the training images. We employ the following algorithm to align the images on the image with the highest resolution (47 MP). First, we detect points of interest. Then, we extract local AKAZE descriptors \cite{alcantarilla2011fast} and, finally, we estimate a homography for every image. Image warping is implemented using bicubic sampling. Note that, while the map computation requires this warping alignment step that may affect the performance, training and prediction can be performed on the original images. We now assume that the training images $\{X_1,\hdots,X_N\}$ are registered.


To estimate the map $M$, we propose to use the network $\phi$ trained in the first stage.
Let $\Psi\in\mathbb{R}^{H',W',C}$ be the feature tensor outputted by the backbone network for a given input image. In our case, since we employ a ResNet-50 network, $\Psi$ corresponds to the tensor before the Global Average Pooling (GAP) layer.
Since Resnet-50 is a fully convolutional network, the dimension $H'$ and $W'$ depends on the input image dimensions, while the number of channels is fixed (i.e.,~$C=2048$). Let $A\in\mathbb{R}^{C}$ and $b\in \mathbb{R}$ be the trained parameters of the final regression layer obtained in the first stage.
The network prediction is given by:

\begin{equation}
   \hat{y} = \sigma \left( A^\top\cdot \text{GAP}(\Psi) + b\right) 
\end{equation}
where $\sigma$ denotes the sigmoid function. While the network returns one single output scalar per input image, we want to obtain one value per pixel location. In order to adapt the class activation map framework \cite{zhou2016learning} to our regression setting, we propose to compute a score for every feature map location $(h,w) \in [1:H'] \times[1:W']$:
\begin{equation}
S_{h,w} = \sigma \left(A^\top\cdot \Psi[h,w]) + b\right)
\end{equation}
where $\Psi[h,w]\in\mathbb{R}^C$ denotes the feature vector at the location $(h,w)$. Note that, the resulting map $S=(S_{h,w})_{(h,w) \in [1:H'] \times[1:W']}$ has a dimension $H'\times W'$ that is different from the initial input image size $H\times W$. This size difference depends on the network architecture. In the case of Resnet-50, we obtain a ratio 32 between the input and the feature map dimensions. Therefore, we resize the score map $S$ to the dimension $H\times W$ using bicubic-sampling. This procedure, is applied to every image of the training set. Thus, we obtain the set of score maps $\{\tilde{S}_1,\hdots,\tilde{S}_N\}$

We propose to define the confidence score map $M$ as the location-wise variance of the score maps $\tilde{S}_n$ over the whole training set. The motivation for this choice is that, discriminating regions have a higher variance than non-discriminating ones. Indeed, we observe that the scores produced by the ConvNet $\phi(\cdot, \theta)$  over non-discriminative regions tend to have small variance. Conversely, on discriminating patches, the networks predict values with a wide range leading to high variance.



\subsection{Final Training and Prediction}
In the last stage of our pipeline, we select the chart region with the highest confidence score value in $M$. 
In our preliminary experiments, we observed that using a region width approximately six times larger than the network input size leads to good performance. In our case, we use a square region of $1200\times1200$ pixels.
In this region, we select random patches that are used as a training set. We re-train the network $\phi$, starting again from ImageNet pre-training weights. 

At test time, assuming an image with an unknown quality score, we extract patches in the selected region. The final score is given by the average over the different patches.

\section{Experiments}
In this section, we perform a thorough experimental evaluation of the proposed pipeline. We implemented our method using Tensorflow and Keras. When training the ConvNets (stages 1 and 3), we employ Adam optimizer following \cite{guideline}, with a starting learning rate of $3\cdot 10^{-3}$ with a decay of 0.1 every 40 epochs for a total of 120 epochs.
In our model, we assume that all the images have the same resolution.
The reason for this choice is that we want the image details to be analyzed at the same scale, as a human observer would do.
In practice, the resolution depends on the device. Therefore,  we preprocess all the training images resizing them to the highest resolution of the dataset using bicubic upsampling. This solution is preferred to downsampling to a common lower resolution since texture quality is not invariant to downsampling. In addition, due to possible lens shading, we remove the sides of the images.


\subsection{Datasets}
\label{sec:datasets}
\begin{figure}[h!]
\centering
\begin{subfigure}[t]{0.15\textwidth}
\includegraphics[width=\textwidth]{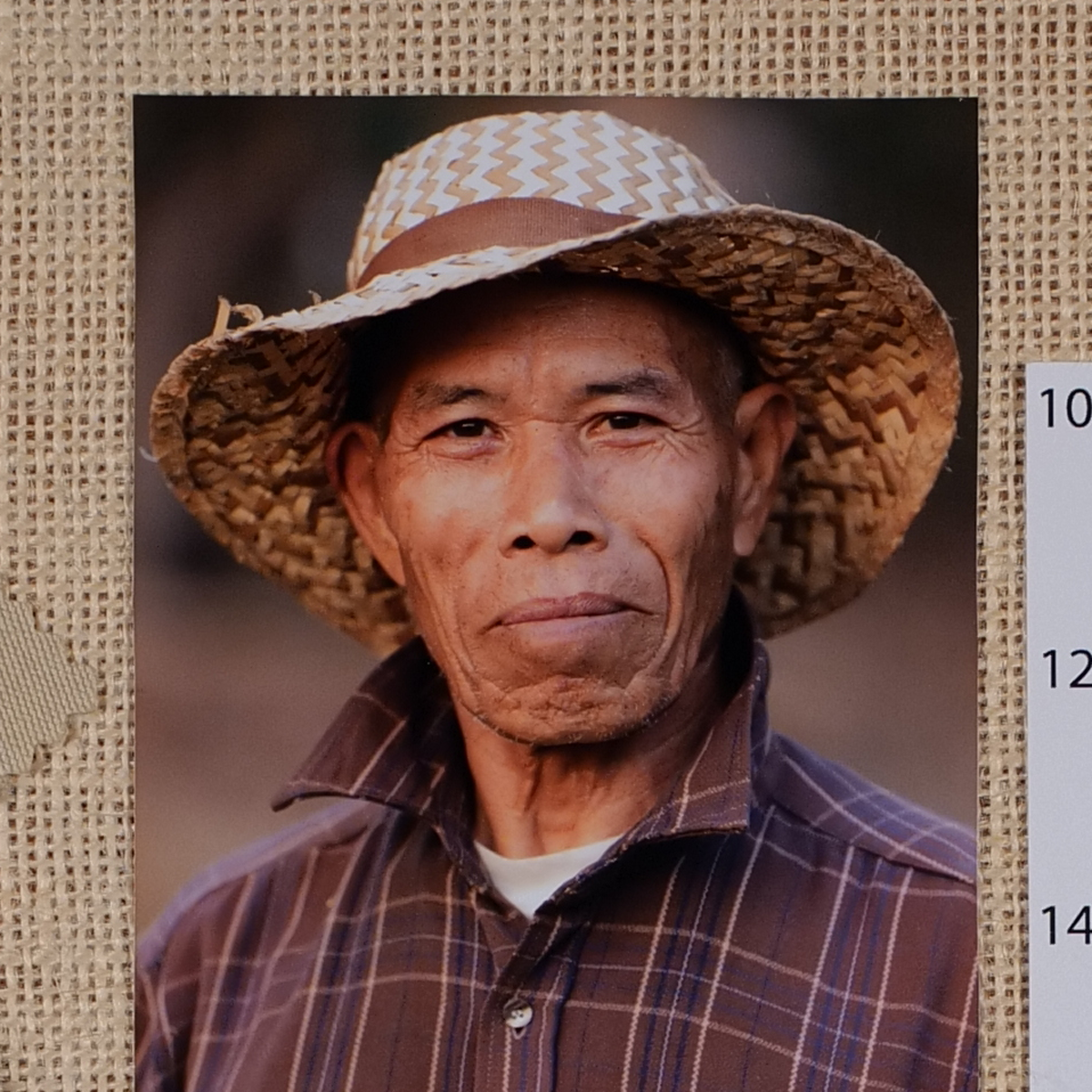}
\end{subfigure}
\begin{subfigure}[t]{0.15\textwidth}

\includegraphics[width=\textwidth]{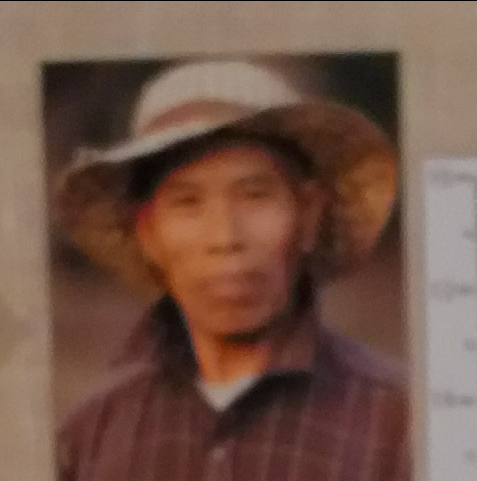}
\end{subfigure}
\begin{subfigure}[t]{0.15\textwidth}

\includegraphics[width=\textwidth]{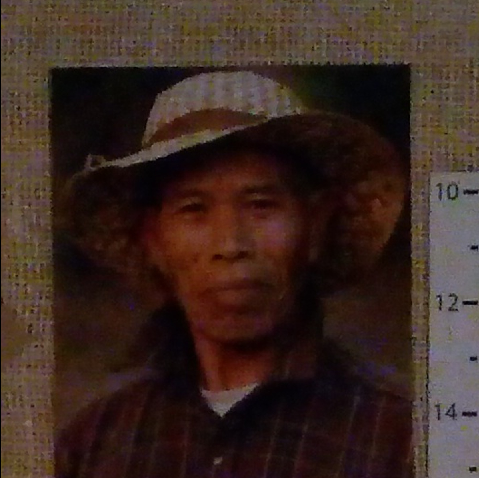}
\end{subfigure}
\caption{Patches of high (left) and low-quality (center and right)  images from our \emph{Still-Life} dataset}
\label{fig:qualityexample}
\end{figure}

\begin{figure*}[h!]
\begin{subfigure}[t]{0.5\textwidth}
\centering
\includegraphics[height=1.8in]{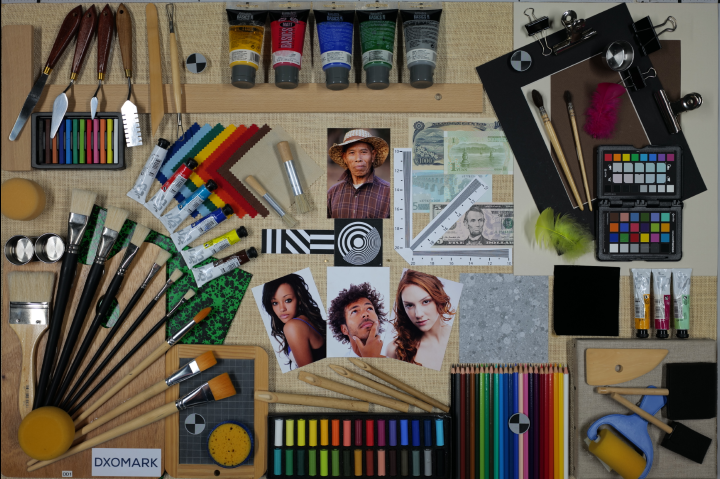}
\caption{\emph{Still-Life} Chart}\label{fig:pChart} 
\end{subfigure}
\begin{subfigure}[t]{0.5\textwidth}
\centering
\includegraphics[height=1.8in]{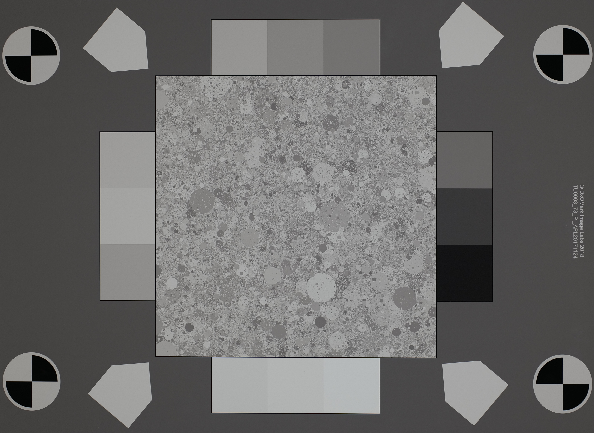}
\caption{\emph{Gray-DL} Chart} 
\end{subfigure}
\caption{Still-Life Chart used in our experiments. The \emph{Still-Life} chart contains many diverse objects with varying colors and textures while the \emph{Gray-DL} chart depicts random gray-scale circles.}
\end{figure*}

\paragraph{Charts and devices}
As there is no well-established reference dataset for our problem, we collected annotated data using three different charts. 
\begin{itemize}
    \item \textbf{Still-Life}: First, we use the chart displayed in Fig.~\ref{fig:pChart}. This dataset is referred to as \emph{Still-Life}. The chart is designed to evaluate several image quality attributes and to present diverse content: Vivid colors for color rendering, fine details, uniform zones, portraits as well as resolution lines and a low-quality Dead-Leaves version. Images are acquired using 140 different smartphones and cameras from different brands commonly available in the consumer market.
    In Fig.~\ref{fig:qualityexample}, we provide an example patch captured using three different cameras. The left image corresponds to a high-quality device while the two others are obtained with low-quality devices. It illustrates the nature of distortions that appear in this dataset with different intensity.   
    To obtain a larger database and predictions robust to lighting conditions, we shoot the chart using five different lighting conditions: 5 lux tungsten, 20 lux tungsten, 100 lux tungsten, 300 lux TL84, 1000 lux D65. Note that process is repeated for every device. 
    \item \textbf{Gray-DL}: Second, we employ the dead-leave chart proposed in \cite{cao2009measuring}. As mentioned in Sec~\ref{DLAcutanceDesc}, this chart depicts gray-scale circles with random radius and locations. In all our experiments, we refer to this dataset as \emph{Gray-DL}. We use the same five lighting conditions and devices as for the \emph{Still-Life} chart. 
    \item \textbf{Color-DL}: Finally, we complete our experiment using the dead-leaves chart proposed in \cite{sumner2017effects}. By opposition to \emph{Gray-DL} this chart is colored (an image can be found at \cite{CCChart}). For this chart, we employed a limited number of devices. More precisely, we employ only 14 devices with the same five lighting conditions as for the other charts. The low number of devices is especially challenging for learning-based approaches.
\end{itemize}

\paragraph{Annotations}

In order to train and evaluate the different methods we need to provide ground-truth annotation for each device. Note that, the annotation must be provided for each pair of device and lighting conditions.
To obtain quality annotations that are a reliable proxy of the perceived visual quality, annotations are provided by human experts. Images to be evaluated were inserted among a fixed set of 42 references, and very high-quality prints were provided to help them judge of the authenticity of the details. Annotators were asked to compare the images using the same field of view for every image, using calibrated high-quality monitors where images are displayed without applying any down-sampling but with a possible digital zoom for the lower resolution image. Each position among the set of references is assigned a score between 0 and 1. 
In the case of the Dead-Leaves charts, since the charts are unnatural images, human perceptual annotation is problematic. 
Therefore, we chose to use the annotations obtained on the \emph{Still-Life} also for the dead-leaves charts, rather than annotating the images. The \emph{Still-Life} chart contains diverse textures similar to what real images would contain. In this way, we obtain a subjective device evaluation in a setting more similar to real-life scenarios. 

\subsection{Metrics}

In our problem, relying on standard classification or regression metrics is not straightforward. Indeed, MTF-based methods predict a quality score that is not directly comparable to the score provided by human annotators. A straightforward alternative could consist in computing the correlation between the predictions and the annotation. However, the underlying assumption that the predictions of each method correlate linearly with our annotations may not hold and bias the evaluation. Therefore, we decided to rely on two distinct metrics based on the correlation of the rank-order. First, we adopt the Spearman Rank-Order Correlation Coefficient (\emph{SROCC}) defined as the linear correlation coefficient of the ranks of predictions and annotations. Second, we report the Kendall Rank-Order Correlation Coefficient (\emph{KROCC}) defined by the difference between concordant and discordant pairs divided by the number of possible pairs. The key advantage of this second metric lies in its robustness to outliers.

For all visual charts, the dataset is split into training and test sets as follows. First, among the devices we use in our experiments several are produced by the same brand. So, to avoid bias between training and test, we impose no brand-overlap between training and test sets. Second, as a consequence of such constrain, a limited number of brands may appear in the test set. To avoid evaluation biases towards specific brands, we use a $k$-fold cross-validation with $k=16$.

In order to measure the impact of the number of devices on the performance, we perform experiments with a variable number of devices. For all the experiments on the \emph{Gray-DL} and \emph{Still-life} charts, we report the results obtained using subsets of size 20, 60, 100 and 140 devices. For a given number of devices, each experience is performed over the same devices set. Note that, for every method, the complete pipeline is repeated independently for every subset. 
\subsection{Ablation study}
\label{sec:ablation}
In order to experimentally justify our proposed method, we compare three different versions of our model:
\begin{itemize}
    \item \emph{Random Patch}: In this approach, random patches are selected from the whole chart at both training and testing time. 
    \item \emph{Random Region}: We then restrict the random patch extraction to a single zone, chosen randomly. We report the average over five random regions.
    \item \emph{Selected Region}: In this model we employ our full pipeline as described in Sec.~\ref{sec:method}. In particular, training and test are performed using the selected region.
\end{itemize}
In these three models, we employ a ResNet-50 backbone trained using the same optimization hyper-parameters. 

\begin{table*}[t]
\begin{center}
\caption{Ablation Study: we measure the impact of region selection comparing three baseline models on the \emph{Still-life} chart. \emph{SROCC} and \emph{KROCC} metrics are reported.}
\begin{tabular}{ cccccccccc}
\toprule
 \textbf{Number of devices} & 20 & 60 & 100 & 140 && 20 & 60 & 100 & 140\\ 
 \midrule
 & \multicolumn{4}{c}{\textbf{SROCC}} & \hspace{0.5cm}& \multicolumn{4}{c}{\textbf{KROCC}}\\
  \midrule 
Random Patch  & 0.626 & 0.818 & 0.784 & 0.806 && 0.433 & 0.617& 0.588 & 0.613\\
Random Region & 0.795 & 0.863 &  0.866 & 0.879 && 0.606& 0.680 &0.682 &0.700\\
Selected Region (Full model) & \bf 0.830 & \bf 0.912 &  \bf 0.890 & \bf 0.900 && \bf0.638 & \bf0.740 &\bf0.716 & \bf 0.728\\
 \bottomrule
\end{tabular}
\label{ablation}
\end{center}
\end{table*}

The results obtained on the \emph{Still-life} chart are reported in Table.\ref{ablation}.  First, when using the \emph{random patches} variant, the model trained on 20 devices performs poorly both in terms of \emph{SROCC} and \emph{KROCC} compared to other variants. In this case, we see that it is required to dispose of at least 60 devices to get satisfying performances. Second, we observe that restricting random patches extraction to a region randomly selected leads to better performance than if we do not restrict to this region. The gain is visible for every number of devices and for both metrics. It may be explained that the decreased diversity in content leads to a ConvNet that is specialized in a specific region of the chart. In other words, the benefit of a more restrained input diversity is larger than the benefit of a larger and more diverse training set. 
Finally, our full model reaches the best performance for both metrics and for every number of devices. This better performance independently of the training sub-set demonstrates the robustness of the proposed method. Interestingly, we obtain performances with 20 devices similar to the performance of the \emph{Random Patch} model with 140 devices. Overall, this ablation study illustrates the benefit of selecting specific regions for texture quality measurement.


\subsection{Qualitative analysis of our region selection}
\begin{figure}[h!]
\centering
\includegraphics[width = 0.45\textwidth]{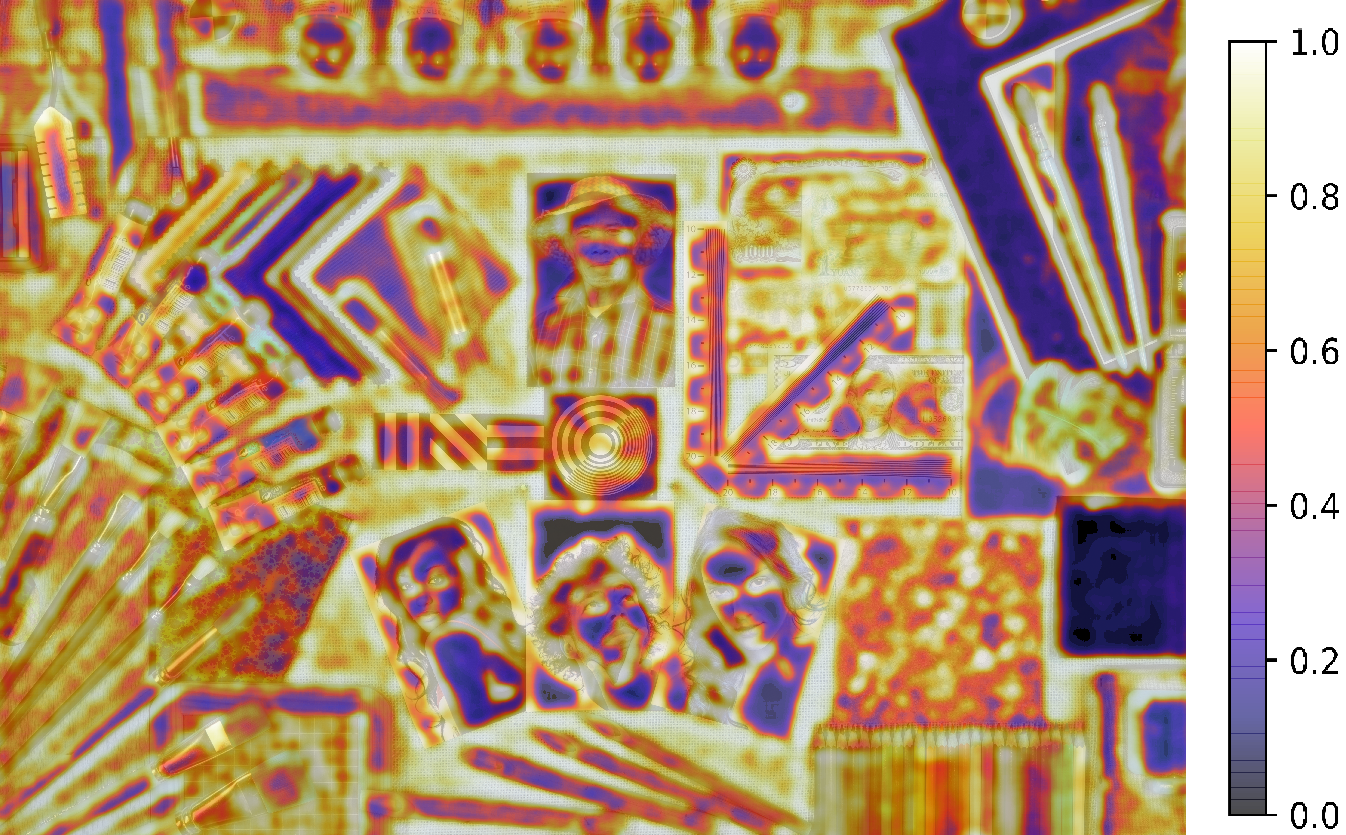}
\caption{Normalized discriminant-region map $S$ (better viewed with a digital zoom). For display, we employ histogram equalization for normalization and obtain values from 0 to 1.}
\label{themap}
\end{figure}

In order to further study the outcome of our region selection algorithm, we display the resulting map (Fig.~\ref{themap}) of relevant zones.

We observe that uniform regions are considered by our algorithm as the least discriminant for texture quality assessment. In particular, this is visible in the bottom-right regions on the black square patch. On the contrary, regions with low contrast and many small details appear to be more discriminant (see around the banknote region). Results on wooden regions seem to depend on wood grain.  

\begin{figure}[h!]
\centering
\begin{subfigure}[t]{0.17\textwidth}
\includegraphics[width=\textwidth]{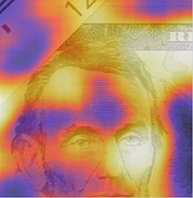}
\end{subfigure}
\hspace{1mm}
\begin{subfigure}[t]{0.17\textwidth}

\includegraphics[width=\textwidth]{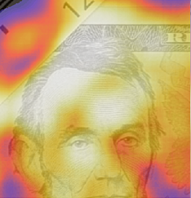}
\end{subfigure}

\bigskip
\begin{subfigure}[b]{0.17\textwidth}

\includegraphics[width=\textwidth]{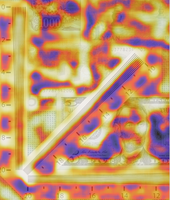}
\caption{Low quality images} 
\end{subfigure}
\hspace{1mm}
\begin{subfigure}[b]{0.17\textwidth}

\includegraphics[width=\textwidth]{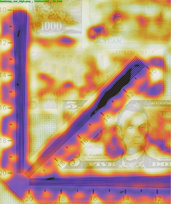}
\caption{High quality images} 
\end{subfigure}
\caption{Comparison of discriminant-region maps for high-quality images and low quality. We display two patches extracted from the confidence maps obtained when using texture quality maps only from high (\ie~Left) and low (\ie~Right) quality images.}\label{fig:high_low}
\end{figure}

This analysis is performed considering all the images. We now propose to analyze the regions that discriminate devices among only low quality or only high-quality images. For this analysis, the test set is split to according to the ground-truth score. In this way, we compute two discriminant maps. Two small crops of these two maps are shown in Fig.\ref{fig:high_low}. Interestingly, we observe restricting our analysis to high quality or lower quality images leads to differences in results. For example, we observe that the resolution lines (in the bottom row of \ref{fig:high_low}) discriminate for low-quality images, but not for higher quality images. Conversely, areas exhibiting only very fine details are not the most useful for low-quality images. In particular, the forehead of the man is not discriminant among low-quality images, while this region is highly discriminant among high-quality images. It shows that the region with very fine details are discriminant only among high-quality images since these details are completely distorted by all the low-quality devices. 

\subsection{Comparison to state of the art}
\begin{table*}[ht!]
\begin{center}
\caption{Comparison of deep learning systems on different charts to \cite{cao2009measuring}.\emph{SROCC} and \emph{KROCC} metrics are reported.}
\begin{tabular}{cccccccccccc}
\toprule
 \multicolumn{2}{c}{\textbf{Number of devices}} &\hspace{0.5cm}& 20  & 60 & 100 & 140 &\hspace{0.5cm}& 20 & 60 & 100 & 140\\ 
  \midrule
 \textbf{Method}&\textbf{Chart}&& \multicolumn{4}{c}{\textbf{SROCC}} &&  \multicolumn{4}{c}{\textbf{KROCC}}\\
 \midrule 
 RR Acutance \cite{cao2009measuring} & \emph{Gray-DL} &&\bf 0.704 & 0.794 & 0.747 & 0.788 && \bf 0.533 & 0.595 & 0.592 & 0.592\\
 ResNet \cite{he2016deep} & \emph{Gray-DL} &&  0.641 & 0.795 &  0.792 & \bf 0.824 && 0.464 & 0.598& 0.592& 0.630\\
 \midrule
$DR^2S$ (Ours) & \emph{Still-Life} & & \bf 0.830 & \bf 0.912 & \bf 0.890& \bf 0.900&& \bf 0.638 & \bf 0.740 & \bf 0.716 & \bf 0.728\\
 \bottomrule
\end{tabular}
\label{CompSota1}
\end{center}
\end{table*}

In this section, we compare the performance of our approach to existing methods. This comparison is twofold since both the methods and the charts need to be compared. 
We perform two sets of experiments. In the first set of experiments, we compare different methods on the two large datasets recorded with the \emph{Gray-DL} and the \emph{Still-Life} charts and the same 140 devices. The second set of experiments consists of a comparison of the devices on the \emph{Color-DL} chart. This second set of experiments is highly challenging for learning-based methods because of the limited amount of training data.

\subsubsection*{Large database experiments}

First, in our preliminary experiments, we observed that, for this experience, adding a small amount of Gaussian noise and random change in exposition leads to better performance. This data-augmentation is performed on the fly on every training patch. Our main competitor is the \emph{RR acutance} methods proposed in \cite{cao2009measuring}. For the acutance computation, viewing conditions were set to 120 centimeters printing height and 100 centimeters viewing distance. Note that, the \emph{RR acutance} method is intrinsically designed for the Dead-Leaves charts and cannot be used for the \emph{Still-Life} chart. We include a second deep learning-based method for the \emph{Gray-DL} chart in our comparison. This approach consists of a ResNet-50 \cite{he2016deep} where the classification layer is replaced by 3 additional fully-connected layers and a linear regression layer. For this approach, we employ the \emph{Random patch} strategy described in Sec. \ref{sec:ablation} inside of the texture region. Importantly, we do not report the performance of \emph{DR$^2$S} on the \emph{Gray-DL} chart since the chart is designed to be uniformly discriminant for texture quality assessment.

Quantitative results are reported in Table.~\ref{CompSota1}. First, we observe that with a limited number of devices for training (e.g. 20 devices), \emph{RR Acutance} performs better than ResNet-50. However, the proposed approach clearly outperforms the texture-MTF based method ($+0.126$ and $+0.105$ in \emph{SROCC} and \emph{KROCC}, respectively). It shows that when few training samples are available, selecting the appropriate regions is essential for good performance. ResNet performance increases with the number of devices: with 140 devices, ResNet-50 clearly outperforms \emph{RR Acutance} according to both metrics showing the potential of learning-based methods. While comparisons between results obtained using different charts must be interpreted with care, this result clearly shows that a learning-based approach can be intrinsically better than acutance-based methods using the exact same input images. 
Finally our \emph{DR$^2$S} method on the \emph{Still-life} chart leads to the best results according to both metrics and for every number of devices.

\subsubsection*{Small database experiments}

Concerning the second set of experiments, we compared the different methods on the \emph{Color-DL} chart. Note that the 14 devices of the \emph{Color-DL} chart are a subset of the devices of the \emph{Gray-DL} and \emph{Still-Life} charts. Consequently, we also performed experiments using only these 14 devices on these two other charts. Note that this setting is very challenging for the two learning-based methods (ResNet and \emph{DR$^2$S}) because of the limited amount of training data. Therefore, for the two learning methods, we perform two experiments. First, training and test are performed using 14-fold cross-validation on the exact same data as the other methods. Second, we train our model on the complete database and test on the 14 devices in common with the \emph{Color-DL} chart. These two variants are referred to as \emph{Restricted} and \emph{Full}. 
Again, we do not report the performance of \emph{DR$^2$S} on the \emph{Gray-DL} and \emph{Color-DL} charts since the chart is designed to be uniformly discriminant. Results are reported in Table \ref{CompSota2}.

\begin{table}[t]
\begin{center}
\caption{State of the art comparison : Performance on the 14 devices database. Deep learning systems on perceptual and Gray-DL are compared to \cite{cao2009measuring} and \cite{sumner2017effects}}
\begin{tabular}{ ccccccc } 
\toprule
\textbf{Method} & \textbf{Chart} &\hspace{0.5cm}& \textbf{SROCC}& \textbf{KROCC}\\
  \midrule 
\emph{FR Acutance} \cite{sumner2017effects} & \emph{Color-DL} &&0.701 &0.544\\
\midrule
\emph{RR Acutance} \cite{cao2009measuring} & \emph{Gray-DL} &&0.714 & 0.552\\
ResNet - Restricted  & \emph{Gray-DL} &&0.640 & 0.463\\
ResNet - Full  & \emph{Gray-DL} &&0.780 & 0.598\\
\midrule
$DR^2S$ - Restricted &\emph{Still-Life} && 0.746 & 0.569\\

$DR^2S$ - Full  &\emph{Still-Life} &&\bf 0.873 & \bf 0.702\\ 
 \bottomrule
\end{tabular}
\label{CompSota2}
\end{center}
\end{table}

First, we observe that the MTF-based methods perform similarly on the color and gray-scale dead leave charts. It shows that better performance of the proposed model on the \emph{Still-Life} chart is not due to the lack of colors in \emph{Gray-DL} but to its content. Second, using the restricted database, both learning-based methods, ResNet and $DR^2S$, under-perform MTF-based predictions. However, when the amount of training data is sufficient, both methods outperform \emph{FR Acutance} and \emph{RR Acutance}.

\section{Conclusion and Future Work}

In this paper, we proposed DR$^2$S, a method which learns to estimate a perceptual quality score. To this end, our algorithm selects the chart region that is the most suitable for texture quality assessment. Our results also suggest that, if enough training samples are available, learning-based methods outperform MTF-based methods.
A limitation of our method is that we select only a single region. However, texture quality is known to be multi-dimensional. Consequently, as future work, we plan to extend our method to multiple regions in order to highlight several complementary discriminant features and better measure the intrinsic qualities of a device.



{
\bibliographystyle{IEEEtran}
\bibliography{bibfile}
}

\end{document}